# Yes, DLGM! A novel hierarchical model for hazard classification

**Zhenhua Wang[1], Ming Ren[1]\*, Dong Gao[2], Bin Wang[2]**

[1]School of information resource management, Renmin University of China, Beijing, 100029, China

[2]College of Information Science and Technology, Beijing University of Chemical Technology, Beijing, 10029, China

zhenhua.wang@ruc.edu.cn; renm@ruc.edu.cn; gaodong@mail.buct.edu.cn; 2019210478@mail.buct.edu.cn

*Abstract*:

Hazards can be exposed by HAZOP as text information, and studying their classification is of great significance to the development of industrial informatics, which is conducive to safety early warning, decision support, policy evaluation, etc. However, there is no research on this important field at present.

In this paper, we propose a novel model termed DLGM via deep learning for hazard classification. Specifically, first, we leverage BERT to vectorize the hazard and treat it as a type of time series (HTS). Secondly, we build a grey model FSGM(1, 1) to model it, and get the grey guidance in the sense of the structural parameters. Finally, we design a hierarchical-feature fusion neural network (HFFNN) to investigate the HTS with grey guidance (HTSGG) from three themes, where, HFFNN is a hierarchical structure with four types of modules: two feature encoders, a gating mechanism, and a deepening mechanism.

We take 18 industrial processes as application cases and launch a series of experiments. The experimental results prove that DLGM has promising aptitudes for hazard classification and that FSGM(1, 1) and HFFNN are effective. We hope our research can contribute added value and support to the daily practice in industrial safety.

*Keywords*:

hazard classification; grey model; hierarchical feature fusion neural network; deep learning; industrial process; HAZOP.

### *Nomenclature*

Notations and Observations

| | |
|---|---|
| HAZOP | Hazard and operability analysis |
| DLGM | The proposed hazard classifier via deep learning under the guidance of grey model |
| GM(1, 1) | Grey model composed of first order differential equation with single variable |
| HFFNN | The proposed hierarchical feature fusion neural network |
| HTS | Hazard time series |
| HTSGG | Hazard time series with grey guidance |
| FSGM(1, 1) | GM(1, 1) with Fourier series |
| BERT | Bidirectional Encoder Representations from Transformers |
| CNN | Convolutional neural network |
| BiLSTM | Bidirectional long short-term memory |
| FC | Fully connected neural network |

SLFE         Sentence-level feature encoder

LLFE         Multi-local feature encoder

GAME       gating mechanism

SDM         superposition deepening mechanism

## 1. INTRODUCTION

The increasing demand of people has enabled the rapid development of modernization and industrialization, more and more industrial processes have sprung up, such as oil refining process and natural gas exploitation process. These heavy industries often face a large number of hazards in the production, dispatching and other operation processes, especially those involving critical equipment and materials, which are usually in extreme environments such as high temperature and ultra-high pressure. Once unprepared hazards occur, the enterprise will be dealt a disastrous blow, the daily activities of the masses will be restricted, and even the national volume and the revitalization of industrial production will be affected [7-9]. Indisputably, guarding against hazards is a top priority.

Fortunately, hazard and operability analysis (HAZOP) can effectively resist and prevent hazards for almost any type of industry and system [10, 36-43]. HAZOP is a leading paradigm in which enterprises and expert teams work together to conduct hazard analysis and prevention for the process in the form of brainstorming. It starts with each node in the process (such as clean sour gas knockout drum) and its deviation (such as low liquid level), analyzes the hazards triggered by them, and gives appropriate suggestions and measures [10-15].

In China, before each process is released or puts into production, the enterprise must unite the expert group to complete HAZOP on it [1]. The Measures for the Administration of Emergency Management Standardization [2] and the Catalogue of Safety Classification and Management of Hazardous Chemical Enterprises (2020) [3] issued by the Chinese government once again emphasize this policy and endow HAZOP with indispensability and obligation. In China's journey towards socialism with Chinese characteristics in the new era, a large number of relevant policies were issued [4], which established the status of HAZOP and also reflected the necessity and urgency of hazard classification.

Hazard classification is critical to policy assessment, decision support and safety early warning, and has a great deal of practical application value [5, 16-19, 23-25]. It can promote enterprises to understand and manage the process, provide unique and reasonable reference for relevant assessment tasks. Specifically, it is beneficial to employees' awareness of hazards, daily maintenance and scheduling, and is convenient for engineers to master the whole process operation and deal with hazards efficiently and calmly. For the expert team, hazard classification can provide decision support and can further brighten the progress of HAZOP. However, there is no research on this important field at present.

In view of the above, in this paper, we present a novel model termed DLGM via deep learning under the guidance of grey model to classify the hazards analyzed by HAZOP. DLGM studies the three themes for measuring hazard, namely, the severity theme, the possibility theme and the risk theme with their respective levels. To our knowledge, there are no relevant works dedicate to the comprehensive research on these three themes, and our work bears the brunt.

The prosperity of deep learning in exploring classification problems [26-31] provides an excellent support for this research, thus, deep learning algorithm lays the foundation for DLGM. Furthermore, we perceive that the hazard strictly follows the propagation path that changes over time, since it flows straight according to the logic analyzed, which also conforms to its evolution in real life. Therefore, DLGM can treat hazards creatively from the perspective of time series. Due to the brilliance of the grey model in analyzing time series problems, such as the typical GM(1, 1) [32-35], we can take the structural parameters of the grey model as a class of guidance or feature, since we believe that the structural parameters corresponding to the hazards at the same level may have potential correlations, which can guide the deep learning model to distinguish different hazards. More noteworthy, in order to alleviate the obvious fluctuation of the hazard time series, we introduce Fourier series to approximate this forcing.

Formally, DLGM enjoys three programs: hazard vectorization, grey guidance and feature processing. First, we operate the pre-training language model BERT to transform hazard into hazard vector. Secondly, we regard the hazard vector as a class of time series, build a GM(1, 1) with Fourier series (FSGM(1, 1)) to model it, and get the grey guidance in the sense of structural parameters. Finally, we design a novel hierarchical feature fusion neural network (HFFNN) to investigate hazard time series with grey guidance (HTSGG). Where, HFFNN is a hierarchical structure with four types of modules: two feature encoder modules for capturing the sentence-level features and multi-local features of HTSGG, a feature gating mechanism module for fusing the captured features, and a deepening mechanism module for integrating the three to predict the level of hazard under three themes.

We hold 18 processes as application cases, such as 800 $m^3$ / h natural gas hydrogen production process and 4 million T / a coal indirect liquefaction process, etc. On this basis, we launch a series of experiments. The competition experiment implies the effectiveness and gratification of DLGM, and the ablation experiment reflects the feasibility of the guidance of FSGM(1, 1), and the advancement of HFFNN. We hope that our work can supplement benefits to the daily operation and progress of industrial safety, inspire other dedicated researchers, and stabilize the orderly development of national industry. The main contributions are as follows:

1. We present a novel model termed DLGM to classify hazards analyzed by HAZOP from three themes.

2. We construct a GM(1, 1) with Fourier series (FSGM(1, 1)) to guide the hazard classification.

3. We conceive a novel hierarchical feature fusion neural network (HFFNN) to capture and fuse features.

4. We integrate 18 specific industrial processes to support application case studies.

5. A series of experiments demonstrate the effectiveness of DLGM, FSGM(1, 1) and HFFNN.

We review the hazard in HAZOP and the classification investigation in Section 2, and illustrate our DLGM classifier in Section 3. In Section 4, we present information about the industrial processes in the application cases. Section 5 records the experiments and their results analysis. The discussion and conclusion are stated in Section 6 and Section 7 respectively.

## 2. RELATED WORK

The cognition of hazard and classification investigation can deepen the approval of the significance of our work.

### 2.1. Cognition of hazard

HAZOP is an analytical methodology and engineering technique with strong generalization and inclusiveness led by expert teams. It can make safety inference for almost any type of industry and system by detecting the hazards [10], such as Brazilian waste pickers' cooperatives [36], sustainable and renewable palm oil industry [37], biomass supply chain optimization [38] and China fusion engineering test reactor (CFETR) central solenoid model coil (CSMC) heat treatment system [39], etc. [40-43]. Therefore, the hazards analyzed by HAZOP planning are diverse, typical and universal, with reality and authority, and the contents it covers are very valuable.

Wang et al. [10] revealed that the hazard satisfies

$$IC \rightarrow D \rightarrow \{ME_1, ME_2, ..., ME_i\} \rightarrow C \rightarrow S$$

causality, where, $IC$, $D$, $C$ and $S$ denote the cause, the deviation, the consequence and the suggestion / measure, respectively. ME refers to the middle event, which is often chain reactions of multiple sub middle events. This kind of causality is consistent with the evolution of hazard in real life, that is, the consequences of a hazard being triggered from its origin to its eventual occurrence, and then attached with suggestions. Thus, the hazard follows the propagation over time, and can be regarded as a kind time series.

For example, such a hazard in the electrolytic hydrogen production unit (*Hazard#1*):

LIC1003 circuit fails, LV1003 opening increases, the liquid level of No.21 hydrogen separator is high, the pressure difference between hydrogen and oxygen separators is increased, and the pressure difference on both sides of the inner membrane of the electrolytic cell is increased, resulting in membrane damage and explosion; During the pressure drop of the hydrogen separator, when the pressure is too much lower than that of the oxygen separator, oxygen will flow into the hydrogen separator from the discharge pipeline, which leads to economic losses and casualties, damage to the reputation of the enterprise and no environmental impact, resulting in explosion, which leads to economic losses and casualties, damage to the reputation of the enterprise and no environmental impact. Implement LAL1001 oxygen separator low level alarm and LSLL1001 oxygen separator low level interlock shutdown (BPCS and LAL1001 share the same circuit). It is suggested that the alarm, control and interlock functions should be separated and independent to prevent all the liquid level protection measures from failing at the same time.

There are three thematic directions for hazard: one is the severity theme measured by five levels under the assessment of property loss, personal injury, environmental pollution and so on, the other is the possibility theme measured by five levels under the assessment of frequency, and the last is the risk theme measured by four levels under the assessment of actions, improvements and suggestions. Take the *Hazard#1* as an example, its levels under the three themes are 3, 5 and 3 in turn.

Please note that the results of HAZOP are recorded in HAZOP report, protected by property rights and kept confidential. To some extent, our work has also brought progress to the transfer and sharing of knowledge.

### 2.2. Classification investigation

Related to this paper is the study of Feng et al. [6], which applies the BERT, BiLSTM and Attention model to HAZOP of diesel hydrogenation, residue hydrogenation and pre-hydrogenation units to complete the severity classification of consequences. However, a considerable gap between it and our paper is that our research is more superior in terms of subject scope, motivation, method, idea and novelty. For example, the object of this paper is the whole hazard rather than the consequence. The scope of this research includes three themes, not just one. Besides, we explore the hazard in a novel and ideological perspective, rather than just calling existing deep learning networks.

Hazard classification is a research field with strong desirability. Even for a specific scene or process, it also has strong practical significance [5, 23-25]. Well, Gordon et al., studied a complete hazard classification system of electric shock to protect the health of construction workers [16]. Ouyang et al., proposed a hazard classification scheme for the general assembly process of automotive spark plugs to achieve accurate processing and management of product fault data [17]. Jahani et al., constructed a hazard classification model based on the fault of platanus orientalis as an environmental decision support system which can reduce the risk of tree collapse [18]. Fang et al., studied the classification of near misses in the construction industry to understand how to mitigate and control the hazards on the construction site [19]., etc. These cases also indirectly prove the importance and practicality of our work. In addition, it is undeniable that deep learning algorithms such as embedding representation and neural network provide a superb basis for these hazard classifications.

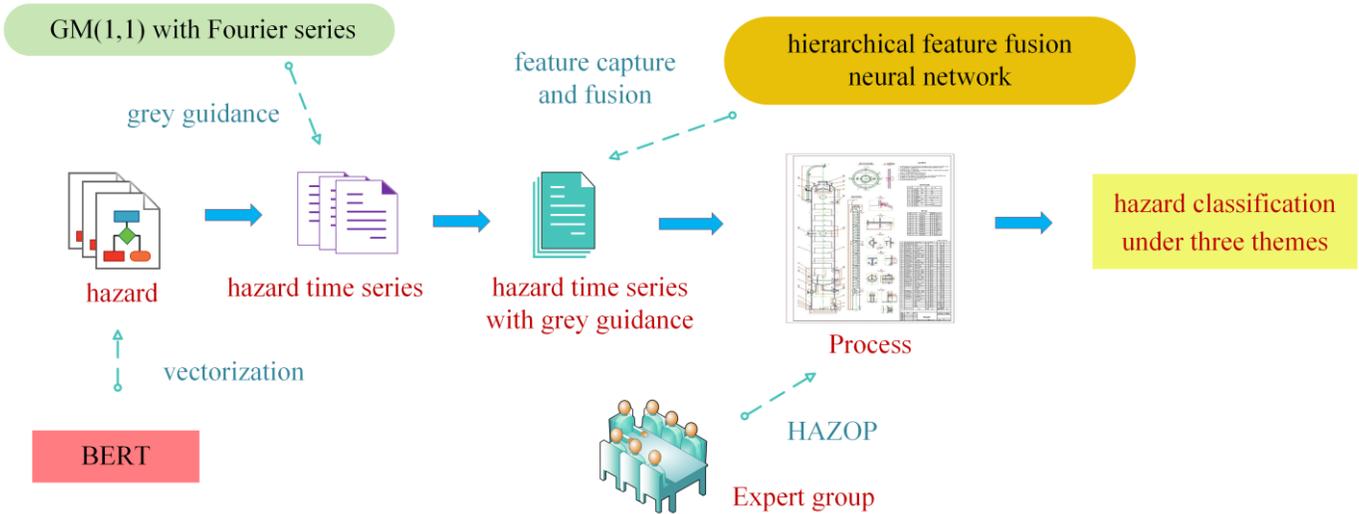

Fig.1: Execution procedure of DLGM.

## 3. METHODOLOGY

This section expounds the proposed DLGM, which is a deep learning framework composed of a hazard vectorization model, an appropriate grey model with Fourier series (FSGM(1, 1)) and a novel hierarchical feature fusion neural network (HFFNN), see Fig.1. First, we employ the pre-training language model BERT to vectorize the pre-treated hazards. Secondly, we regard the hazard vector as a class of time series, build the FSGM(1, 1) to model it, and get the grey guidance in the sense of the structural parameters. Finally, we design the HFFNN to investigate the hazard time series with grey guidance to achieve the classification of hazards under different themes. Details are as follows.

### 3.1. Hazard Vectorization

What this section can engage is the vectorization procedure of the hazard. The input is hazard in text format and the output is hazard vector, as illustrated in Fig.2.

We perceive that the meaning behind the hazard is related to the role played by the equipment and material. For a hazard, the equipment involved acts as 1) the node of the deviation source, 2) the promoter of the middle event, 3) the prevention tool in the proposal, or 4) the victim and the implicated in the consequences, which leads to the hazard belonging to different levels under different themes. Similarly, the role of the material also affects the final determination of hazards. Another thing to consider is that hazards contain complicated terminologies and components of different reactions (such as *Hazard#1*), moreover, there are also differences in their language styles, as well as the wording and phrasing.

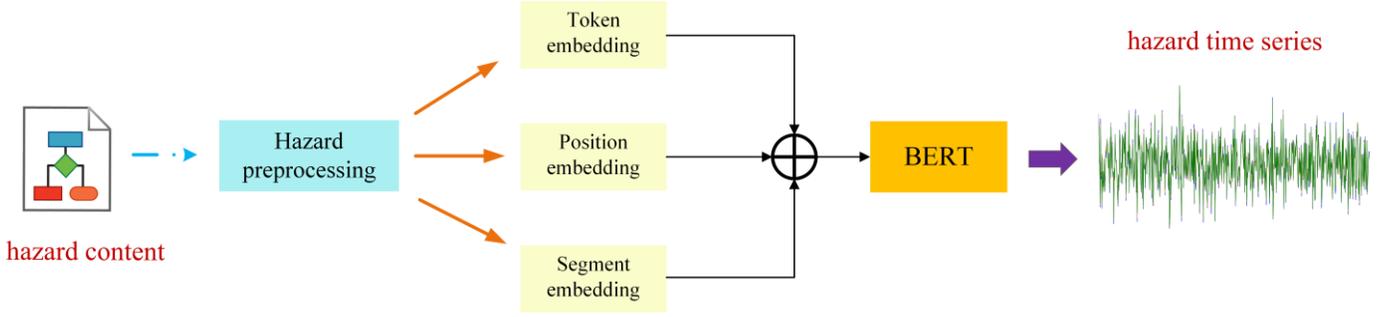

Fig.2: Procedure of Hazard vectorization procedure.

Therefore, we can manipulate BERT, a pre-trained language model that flourishes industrial and social language processing and understanding, to vectorize the hazard [44-47]. Considering that BERT is trained from general corpora via *Transformer*'s encoder components, we fine-tune BERT through hazard corpus to inject it with a priori knowledge in the field of industrial safety.

Specifically, given a pre-cleaned hazard content $W = \{w_1, w_2, ..., w_m\}$, $w_i$ represents the word and $m$ denotes the length of the content. We attach a "CLS" mark to the beginning of $W$ to establish the boundary of the hazard, that is:

$W = \{CLS, w_1, w_2, ..., w_m\}$.

Next, we map $W$ to a representation (see Equ.1) of the sum of the token embedding, the segment embedding and the position embedding, where, the token embedding is formed by the greedy longest-match-first algorithm that searches from Bert's own vocabulary for tokens processed by the WordPiece algorithm, the segment embedding is formed by sequentially assigning 0 and 1 to adjacent sentences, the position embedding is generated by the sine / cosine formula with different frequencies in different dimensions. We take the concatenation $E_W$ as inputs and straightforwardly feed them to the decoder group of *Transformer* to generate the hazard vector whose dimension of each word is 768 through the mask language operation and the next sentence prediction operation.

$$E_W = E_{token} + E_{segment} + E_{position} \qquad (1)$$

It is worth noting that the represented hazard vector is mapped by the elements in the hazard dynamically in various contexts, and thus additionally contains potential semantic information.

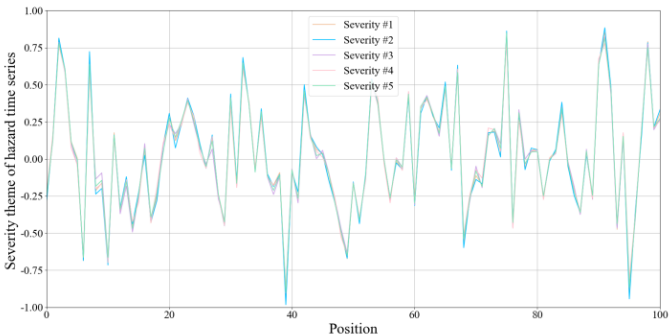

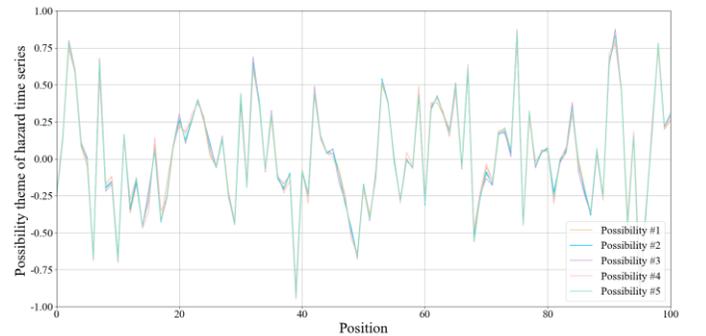

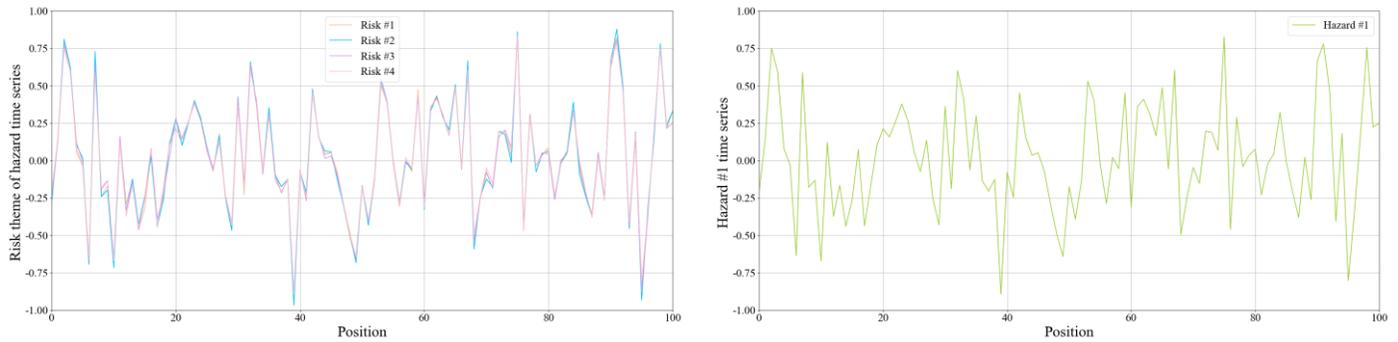

Fig.3: Local manifestation of (1): different levels of random hazard time series under the severity theme, possibility theme and risk theme. (2) *Hazard#1* time series.

We regard the hazard vector as a class time series termed HTS. Fig.3-(1) reflects the local profiles of the mean squeezed HTS we randomly selected about the three themes, respectively, note that these 17 hazards involved vary. Fig.3-(2) is the profile of the *Hazard#1* in Section 2.1. Obviously, all levels under the three themes show quasi-periodic fluctuations, which is close to the noise data whose regularity is hard to explore. It can be realized that the differences in the trends of various hazards among the themes are subtle, which indicates that the hazards do not exist alone, and are related at least in the trend. This is in line with the real situation and also reflects the challenge of our work.

When the perspective is switched to the inside of the respective themes, different colors show different levels of dominance, and the differences between them are of course subtle. We can notice that in the severity theme and the risk theme, the value amplitude of level#2 is relatively more obvious, and the trend is larger, which is consistent with the status of the industrial system in reality, that is, the medium-scale hazards are more likely to be triggered, while in the possibility theme, it is level#3, we believe that in this way, warnings with a slightly higher probability can be paid more attention by the staff, with more precautions reserved. Another real phenomenon is that the amplitude of the highest level of the theme is relatively shallow and narrow, see the severity#5, possibility#5 and risk#4 in the figures, after all, large-sized hazards are rare.

In short, how to better handle HTS is worth pondering. Inspired by the brilliance of grey models in dealing with time series problems, such as the typical GM(1, 1) [32-35, 48], a natural idea is that GM(1, 1) corresponding to different categories of hazards can be used as their respective auxiliary identification, because we believe that each HTS may have its own structural parameters of GM(1, 1), which can be a guide for viewing hazard, similar to the potential features required to meet the deep learning training.

GM(1, 1) enjoys great popularity in the time series with quasi-exponential distribution, but it weakly undertakes the forcing caused by the fluctuation of HTS. In order to alleviate this limitation, we introduce Fourier series into GM(1, 1), termed FSGM(1, 1). Fourier series is a basis function that widely exists in various fields [49-53]. It can extend almost any periodic function, and can provide inspiration for approximating such forcing. The FSGM(1,1) is explained as follows.

### 3.2. GM(1, 1) with Fourier series

This section represents how to embed the Fourier series into GM(1,1) to realize the approximation of the forcing of HTS. The HTS whose hazard is initially characterized is:

$$X^{(0)}(t) = (x_1^{(0)}, x_2^{(0)}, ..., x_n^{(0)}).$$

Its modified first-order cumulative generation series is:

$$X^{(1)}(t) = (x_1^{(1)}, x_2^{(1)}, ..., x_n^{(1)}), \text{ where,}$$

$$x^{(1)}(t) = (\sum_{k=1}^{t} x_k^{(0)} + \sum_{k=1}^{t} x_{k+1}^{(0)}) / 2, t = 1, 2, ..., n$$

The general grey model GM(1, 1) under $X^{(1)}(t)$ is composed of differential equation:

$$\frac{d}{dt} x^{(1)} = \lambda x^{(1)} + \mu \qquad (2)$$

and its difference equation:

$$x_k^{(0)} = \lambda \frac{x_{k-1}^{(1)} + x_k^{(1)}}{2} + \mu. \qquad (3)$$

Where, $\lambda$ and $\mu$ are structural parameters, and $\mu$ represents the forcing term. Equ.3 partly reflects the change of $x^{(1)}$ that satisfies a relationship without additional forcing. The solution of Equ.2 is:

$$x^{(1)} = \eta e^{\lambda t} - \mu / \lambda \qquad (4)$$

Where, $\eta$ is the coefficient related to the initial value, here $\eta = x_1^{(0)} - \mu / \lambda$. It is well known that GM(1, 1) is suitable for $x^{(1)}$ following the power-like trend. However, the HTS shows obvious fluctuation. In order to adapt to the response brought by such impact, we can force $\mu$ to be a periodic function $p(t)$ to accept the solution of Equ.2, that is:

$$\frac{d}{dt} x^{(1)} = \lambda x^{(1)} + p(t).$$

Where, the forcing term $p(t)$ is a periodic function, here we estimate it by Fourier series. In general, the expression of $p(t)$ under Dirichlet condition is:

$$p(t) = \lim_{k \to \infty} [a_0 + \sum_{n=1}^{k} (a_n \cos(n\omega t) + b_n \sin(n\omega t))].$$

Where, $\omega = 2\pi / T$ is the angular frequency, $a_0$, $\{a_n\}$ and $\{b_n\}$ are the Fourier coefficients. Considering that the Fourier coefficients tend to be infinite, we truncate the first $N$ terms ($k = N$) to approximate $p(t)$, that is:

$$p(t) \approx a_0 + \sum_{n=1}^{N} (a_n \cos(n\omega t) + b_n \sin(n\omega t)).$$

Thus, we obtain a GM(1, 1) with Fourier series that can be used for HTS:

$$\frac{d}{dt} x^{(1)} = \lambda x^{(1)} + a_0 + \sum_{n=1}^{N} (a_n \cos(n\omega t) + b_n \sin(n\omega t)). \qquad (5)$$

Where, $\lambda$, $a_0$, $\{a_n\}$ and $\{b_n\}$ are structural parameters that need to be estimated. They are similar to those that can be transferred to the neural network as features to guide hazard classification, which are referred to as "grey guidance" in this paper. $\omega$ corresponds to the number of times that the value of HTS goes back and forth between the positive interval and the negative interval, that is:

$$T = \sum_{i=1}^{n-1} \tau; \quad \tau = \begin{cases} 1, & if\ x_i x_{i+1} < 0 \\ 0, & if\ x_i x_{i+1} > 0 \end{cases}.$$

If there are $T$ such cases, then $\omega = 2\pi / T$. In combination with Equ.4, the analytical solution of differential Equ.5 is:

$$x^{(1)} = \eta e^{\lambda t} + A_0 + \sum_{n=1}^{N} (A_n \cos(n\omega t) + B_n \sin(n\omega t)). \qquad (6)$$

Where, the coefficient $\eta$ related to the initial value is:

$$\eta = e^{-\lambda t_1} (x_1^{(1)} - A_0 - \sum_{n=1}^{N} (A_n \cos(n\omega t_1) + B_n \sin(n\omega t_1))),$$

$A_n$ and $B_n$ are:

$$\begin{bmatrix} A_n \\ B_n \end{bmatrix} = \left\{ \begin{bmatrix} 0 & 0 \\ 0 & 0 \\ 0 & 0 \\ \vdots & \vdots \\ -\lambda & -n\omega \\ n\omega & -\lambda \end{bmatrix}^T \right\}^{-1} \begin{bmatrix} a_0 \\ a_1 \\ b_1 \\ \vdots \\ a_n \\ b_n \end{bmatrix}.$$

These structural parameters can be proved by bringing them into Equ.6.

For Equ.5, considering that the HTS is discrete, we have

$$\frac{d}{dt} x^{(1)} = \frac{\Delta}{\Delta t} x^{(1)}.$$

Note that $\Delta t = (t + 1) - 1 = 1$, and $\Delta x^{(1)} = x_t^{(1)} - x_{t-1}^{(1)} = (x_t^{(0)} + x_{t+1}^{(0)}) / 2$, we have

$$\underbrace{(x_t^{(0)} + x_{t+1}^{(0)}) / 2}_{U} = \underbrace{\lambda x^{(1)} + a_0 + \sum_{n=1}^{N} (a_n \cos(n\omega t) + b_n \sin(n\omega t))}_{D}. \qquad (7)$$

Considering that Equ.7(U) is known and Equ.7(D) contains unknown grey guidance, we use the least square method by minimizing the objective function

$$L(Q) = \| Y - ZQ \|_2^2$$

to estimate Equ.6, where,

$$Y = \begin{bmatrix} x_1^{(0)} + x_2^{(0)} \\ x_2^{(0)} + x_3^{(0)} \\ \vdots \\ x_{n-1}^{(0)} + x_n^{(0)} \end{bmatrix} \qquad Z = \begin{bmatrix} 0.5x_2^{(1)} + x_1^{(1)} & 1 \\ 0.5x_3^{(1)} + x_2^{(1)} & 1 \\ \vdots \\ 0.5x_n^{(1)} + x_{n-1}^{(1)} & 1 \end{bmatrix} \qquad Q = \begin{bmatrix} \lambda \\ p(t) \end{bmatrix}.$$

The estimated value of $Q$ is

$$\hat{Q} = (Z^T Z)^{-1} Z^T Y, \tag{8}$$

which is brought into Equ.6 to obtain the time response function, that is:

$$\hat{x}^{(1)} = \underbrace{\hat{\eta} e^{\hat{\lambda} t}}_{x_t} + \underbrace{\hat{A}_0 + \sum_{n=1}^{N} (\hat{A}_n \cos(n\omega t) + \hat{B}_n \sin(n\omega t))}_{x_f}$$

It is not difficult to find that $x_t$ in the form of exponential function and $x_f$ in the form of Fourier series respectively reflect the trend and fluctuation of the HTS. By adjusting the order of the Fourier series, $x_f$ can represent the fluctuations under different hazards, so it can depict the gray guidance contained in different levels of hazards, and serve to distinguish hazards. Note that this study is not to explore the grey model itself, which is considered a research problem on its own (such as error test) that is beyond our consideration and do not belong to the scope of this paper. Considerately, we will evaluate the feasibility of the GM(1, 1) with Fourier series through ablation experiments.

Therefore, we get that the grey guidance ($GG_{hazard}$) in the sense of structural parameters of HTS is:

$$GG_{hazard} = (\hat{\eta}, \hat{\lambda}, \hat{A}_0, \hat{A}_1, \hat{B}_1, \cdots, \hat{A}_N, \hat{B}_N).$$

It can be regarded as the potential auxiliary identifier like the feature for hazard classification.

In addition, the consideration of the order $N$ of Fourier series cannot be neglected. The smaller $N$ is, the fluctuation of HTS may not be properly fed back. The larger $N$ is, the final classification model may be accompanied with overfitting problem. In order to preliminarily verify this viewpoint, we intend to determine the appropriate $N$ through hazard classification experiments. We leverage a classifier composed of BiLSTM and fully connected neural network (see Section 5.2) to conduct trial experiments on the severity theme, and record $F_1$-score (F1) results, see Table 1 and Fig.3, where "test" refers to the test set and "val" refers to the validation set.

Table 1: The trial results (100%) of $N$

| N | F1 | | N | F1 | |
| | test | val | | test | val |
|---|---|---|---|---|---|
| 1 | 80.92 | 78.88 | 6 | 81.06 | 79.51 |
| 2 | 81.23 | 79.56 | 7 | 80.84 | 79.35 |
| 3 | 81.59 | 79.79 | 8 | 80.70 | 79.38 |
| 4 | 81.37 | 79.74 | 9 | 80.63 | 79.40 |
| 5 | 80.95 | 79.53 | 10 | 80.51 | 79.27 |

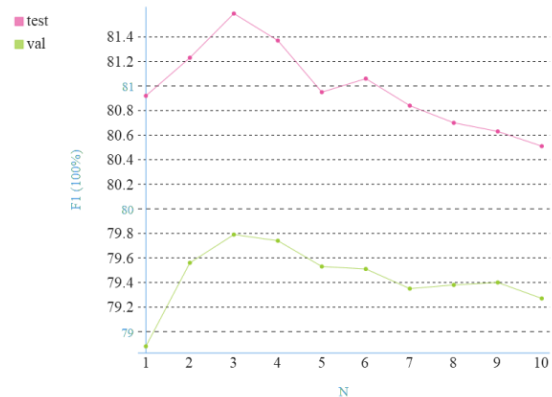

Fig.3: Trial results of the order $N$ of Fourier series.

It can be observed that the performance of the classifier fluctuates with the change of $N$, whether on the test set or on the validation set. When $N < 3$, the performance tends to improve until the performance reaches the best when $N = 3$. Then, when $N$ is greater than 3, the performance gradually fluctuates, but on the whole, the performance shows a decline trend. This is consistent with our view that $N$ is a compromise value, so we set $N = 3$, that is:

$$GG_{hazard|N=3} = (\hat{\eta}, \hat{\lambda}, \hat{A_0}, \hat{A_1}, \hat{B_1}, \hat{A_2}, \hat{B_2}, \hat{A_3}, \hat{B_3})$$

We concatenate the HTS to its $GG_{hazard|N=3}$ grey guidance, see Equ.9, and deliver the formed new sequence information, i.e., HTS with gray guidance (HTSGG), to the hierarchical-feature fusion neural network in the next section for investigation.

$$HTSGG = HTS + GG_{hazard|N=3} \qquad (9)$$

### 3.3. Hierarchical-feature fusion neural network

We conceive a hierarchical feature fusion neural network (HFFNN) to investigate the delivered HTSGG to complete the discrimination of hazards, see Fig.4. HFFNN is a hierarchical structure with three layers and four types of modules: a sentence-level feature encoder (SLFE), a multi-local feature encoder (LLFE), a gating mechanism (GAME) and a superposition deepening mechanism (SDM). First, SLFE and LLFE respectively encode HTSGG to obtain sentence-level features ($f_s$) and multi-local features ($f_l$). Then, the gating mechanism fuses $f_s$ and $f_l$. Finally, SDM deepens the processing of these features by means of module superposition and predicts the level of hazard under three themes. Details are as follows.

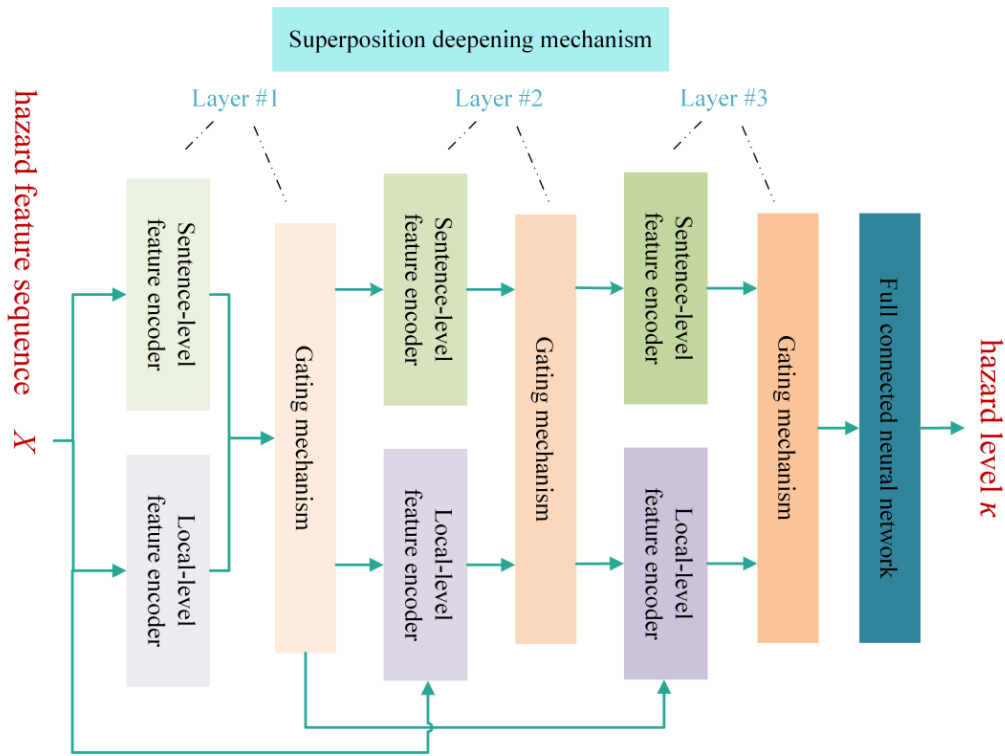

Fig.4: Architecture of hierarchical feature fusion neural network.

### 3.3.1. SLFE

For sequence information, the features between sentences can enhance the performance of the classifier to some extent [55]. Thus, we propose a new sentence-level feature encoder (SLFE) that can capture sentence-level features $f_s$. Considering that the benefits of each element on HTSGG are different, the self-attention mechanism with Gaussian noise [10] is leveraged to focus on each word, and then the HTSGG is converted into $f_s$.

For HTSGG, the hazard feature sequence $X = \{x_1, x_2, \ldots, x_p\}$, we have

$$X = \begin{bmatrix} x_1^1 & x_1^2 & \cdots & x_1^d \\ x_2^1 & x_2^2 & & x_2^d \\ \vdots & & \ddots & \vdots \\ x_p^1 & \cdots & & x_p^d \end{bmatrix}.$$

Where, $p$ is the length of the sequence and $d$ is the dimension of the feature element therein. The self-attention distribution $C$ between the two feature elements is

$$C = softmax(s(X, N)) = \begin{bmatrix} c_1^1 & \cdots & c_1^p \\ \vdots & \ddots & \vdots \\ c_p^1 & \cdots & c_p^p \end{bmatrix}.$$

Where, $N$ is one time Gaussian noise, and $s(X, N)$ is the score function with scaling point product, that is:

$$s(X, N) = (XX^T + N) / \sqrt{d}.$$

The contribution degree $\upsilon_i$ of each feature element mapped to sentence level features is

$$\upsilon = softmax[diag(I - C)].$$

Where, $diag(I - C)$ represents the diagonal elements in the matrix $(I - C)$, that is:

$$diag(I - C) = [1 - c_1^1, 1 - c_2^2, ..., 1 - c_p^p].$$

Thus, the sentence-level feature $f_s$ can be encoded as:

$$f_s = \upsilon X.$$

### 3.3.2. LLFE

This part is the discussion of the proposed multi-local feature encoder (LLFE), We apply multiple CNNs [54] to capture the multi-local feature $f_l$ of HTSGG, which can heighten the representation of the neural network to the hazard and is helpful to the work of the hazard classifier.

For HTSGG, the hazard feature sequence $X = \{x_1, x_2, \ldots, x_p\}$, the output $\mu$ of the convolution layer is

$$u = KDAC(w \cdot x_{i:i+h-1} + b).$$

Where, $KDAC$ is an activation function for text knowledge [56], $w$ is the convolution kernel used to scan the element segment of the sequence, $h$ is the length of the element segment, $b$ is the bias vector, and • is the convolution operation.

The setting of $h$ can directly affect the decision of local features. For example, when $h = 3$, the equipment "费托反应器 (Fischer Tropsch reactor)" will be cut into some invalid fragments, such as "费托反" and "托反应", which makes the corresponding features of the Fischer Tropsch reactor unable to be well preserved and hard to be captured by CNN. To alleviate this dilemma, we respect the style of $h$ to build multiple CNNs, $h$ conforms to the length of key entities in hazards, such as the material and the equipment. We count the length of key entities to set $h$. Considering that the existing entity segmentation tools fail to fit the industrial field, we use the trained named entity recognition model composed of BERT-CRF to extract key entities from hazards [10], see Equ.10, and then we can comprehensively record their lengths. We take 4 million tons / year coal indirect liquefaction process and 1.2 million T / a heavy oil catalytic cracking process as cases (See Section 4 for information about the two processes) to extract material entities (MAT) and equipment entities (EQU). The results are shown in Table 2 and Fig.5, where, Table 2 records the proportions of entities with different lengths $h$, and gives corresponding entity examples.

$$\{EQU, MAT\} = BERT \& CRF(Hazard) \tag{10}$$

Table 2: Statistics of the length of key entities.

| $h$ | Proportion (%) | Entity example |
|---|---|---|
| 2 | 12.34 | 富气 |
| 3 | 9.42 | 除氧水 |
| 4 | 24.13 | 稳定蜡泵 |
| 5 | 35.37 | 液位控制器 |
| 6 | 11.74 | 费托轻质油罐 |
| 7 | 3.27 | 重柴油罐切水器 |
| 8 | 2.15 | 洁净工艺冷凝水罐 |
| 9 | 1.90 | 粉末树脂覆盖过滤器 |
| 10 | 0.68 | 脱吸塔顶气油气分离器 |

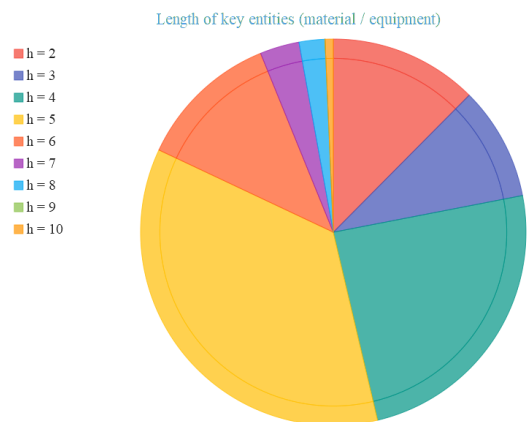

Fig.5: Proportion of key entities about different lengths.

It can be observed that, on the one hand, the lengths of key entities are mainly concentrated on 2, 4, 5 and 6. On the other hand, considering that some equipment with high length often nests materials or equipment with low length ($h$ is generally 2 or 3), such as 粉末树脂覆盖过滤器 (powder resin covered filter) nests 树脂(resin) and 过滤器 (filter). Therefore, the value set of $h$ is $h \in \{2, 3, 4, 5, 6\}$.

Accordingly, the proposed LLFE includes 5 CNNs with different $h$, and the output of the corresponding 5 convolution layers is $\mu_k$, $k = 1, 2, 3, 4, 5$.

Next, in order to avoid overfitting problems and highlight key features, we use *max*-pooling to sample the output value $\mu_k$ of each convolution layer, and obtained multi-local features $f_l$ are

$$\{f_{l \to k}\}_{k=1}^5 = max\text{-}pooling(\{u_k\}_{k=1}^5).$$

### 3.3.3. GAME

This part presents the proposed gating mechanism (GAME). Inspired by Gu et al. [21], the fusion of different types of features is conducive to the performance of classifiers. Thus, we design GAME to filter and fuse $f_s$ and $f_l$ to facilitate hazard classification. See Fig.6.

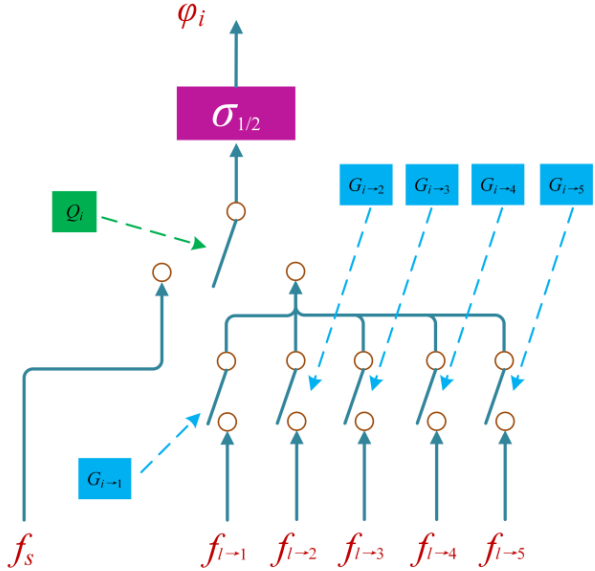

Fig.6: Structure of the proposed gating mechanism.

First, the filter gate mechanism $G_{i \to k}$ is used to control the $f_l$ feature stream output by the previous layer:

$$G_{i \to k} = \sigma(\tau, (W f_{l \to k, i-1} + b)) |_{\tau=0.1}^{0.5},$$

where, $W$ and $b$ are the learned weight and bias respectively, $k$ represents the $k$-th local feature, $\sigma(\tau, x)$ is a homogenized sigmoid function that satisfies

$$\sigma(\tau, x) = sigmoid(\tau x) = 1 / (1 + e^{-\tau x}).$$

We use $G_{i \to k}$ to filter $f_{l \to k, i-1}$, and the processed local feature $r_{i \to k}$ is

$$r_{i \to k} = tanh(W(G_{i \to k} * f_{l \to k, i-1}) + b),$$

where, * denotes element-wise product. We concatenate $r_{i \to k}$ and transfer it to the next part.

Next, we conceive the fusion gate mechanism $Q_i$ meeting

$$Q_i = sigmoid(W f_{s, i-1} + b + V r_i + d)$$

to measure how the transferred feature $r_i$ is fused with the sentence-level feature $f_s$ output from the previous layer. Where, $V$ and $d$ are the learned weight and bias respectively, $r_i$ is the concatenation of $r_{i \to k}$, that is:

$$r_i = concat(\{r_{i \to k}\}, k = 1, 2, 3, 4, 5).$$

The fusion feature $\gamma_i$ is traded by

$$\gamma_i = Q_i * f_{s,i-1} + H_i * r_i,$$

where, $H_i = 1 - Q_i$. This function allows the gating mechanism to treat features from different sources appropriately.

Finally, we modify $\gamma_i$ by

$$\varphi_i = \sigma_{1/2}(W\gamma_i + b)$$

to obtain the output $\varphi_i$ of the gating mechanism, where, $\sigma_{1/2}(\bullet)$ is the $\sigma(\tau, x)$ function under $\tau = 0.5$.

### 3.3.4. SDM

This part describes the proposed superposition deepening mechanism (SDM). The study [57] shows that the representation of deep neural networks can be enhanced by appropriate network superposition. Thus, SDM deepens the final feature representation ability of HFFNN through the superposition of the first three modules, see Fig.4.

Specifically, first, SDM gains sentence-level feature sequence $f_s$ and multi-local feature sequence $f_l$ from the hazard feature sequence $X$, and obtains the fusion feature sequence $\varphi$ of the two through the gating mechanism, see Equ.11.

$$f_s, \{f_{l \to k}\}_{k=1}^5 = SLFE(X), \ MLFE(X)$$
$$\varphi = GAME(f_s, \{f_{l \to k}\}_{k=1}^5) \tag{11}$$

Next, $X$ and $\varphi$ are concatenated to form the feature sequence $X'$. $\varphi$ and $X'$ are converted into new $f_s$ and $f_l$ by SLFE and LLFE, respectively, and the two are fused by GAME to form a new fusion sequence $\varphi'$, see Equ.12.

$$X' = concat(X, \varphi)$$
$$\varphi' = GAME(SLFE(\varphi), MLFE(X')) \tag{12}$$

Then, SDM superimposes the feature sequences calculated by Equ.12 in the form of Equ.13 to obtain the final deepened feature sequence $\varphi''$.

$$\varphi'' = GAME(SLFE(\varphi'), MLFE(concat(X', \varphi'))) \tag{13}$$

Finally, see Equ.14, we pass $\varphi''$ to the full connected neural network and map it to the level space of the hazard through linear layer transformation. Further, we employ *softmax* function to determine the specific level $\kappa$ under different themes.

$$p(y = \kappa \mid \varphi'') = \frac{exp(W_\kappa \varphi'' + b_\kappa)}{\sum_{\kappa_1=1}^{\kappa} exp(W_{\kappa_1}\varphi'' + b_{\kappa_1})} \tag{14}$$

## 4. APPLICATION CASE

We cooperate with 7 enterprises including Shenhua Group and Yanshan Petrochemical to launch HAZOP on 18 specific processes and take them as application cases in this paper. These processes are of great significance in resource utilization, non-renewable resource substitution, environmental protection, sustainable energy development, pollution treatment, industrial and daily necessities, etc. The brief introduction is as follows.

1. 2.2 million T / a diesel hydrofining process: It consists of 14 nodes and reaction units. It uses straight run oil from atmospheric and vacuum distillation unit and coking-oil from coking unit as main raw materials to produce high-quality ultra-low sulfur refined diesel, naphtha and sulfur-rich gas.

2. 300 T / h solvent regeneration process: It is a process of sulfur recovery and utilization in petroleum refining, involving rich and poor liquid ring heat and rich liquid flash system. It relies on chemical absorption and uses alcohol amine to remove acid gas, and then regenerates the rich solution containing hydrogen sulfide through stripping.

3. 1.2 million T / a heavy oil catalytic cracking process: It uses heavy oil such as wax oil, low-pressure wax oil mixed with residue, de-asphalted oil and hydro-demetallized residue as raw materials, and uses molecular sieve catalyst, riser continuous reaction regeneration cycle system and passivator to mainly produce gas raw materials such as high-octane gasoline fraction and light diesel fraction.

4. 600 thousand T / a naphtha isomerization process: The light naphtha from hydrocracking passes through the raw material coalescer to remove the residual water from the upstream alkali washing, and then is mixed with the straight run light naphtha from the aromatic's unit, the light raffinate from the top of the raffinate tower and the reformed pentane. One part of the product is condensed as reflux, and the other part is sent to the feed buffer tank of the depentanizer. The isomerization reaction part adopts the recycle hydrogen process.

5. 120 thousand T / a sulfur recovery process (from Brunei Hengyi Enterprise): It mainly includes two series of Claus sulfur recovery unit and tail gas treatment unit, one series of solvent absorption and regeneration unit, tail gas treatment unit, sulfur forming unit, steam and condensate system unit. The Claus tail gas is collected by the trap and then enters the tail gas treatment unit for purification and recovery.

6. 100 thousand T / a sulfur recovery process (from Sichuan Petrochemical Enterprise): The systems involved include: the procedure of clean acid gas entering the unit and condensate separation, the procedure of ammonia containing acid gas entering the unit and condensate separation, the procedure of sulfur production conversion and process gas heat exchange, the sulfur production furnace, steam generation superheater, tail gas liquid separation tank, tail gas heating, tail gas hydrogenation, tail gas cooling and quench water system, tail gas absorption system, liquid sulfur degassing system, and sulfur forming system.

7. 30 thousand T / a desulfurization and sulfur recovery process: It includes 12 systems that can remove sulfur-containing compounds (such as hydrogen sulfide, hydrogen rich dry gas, catalytic reforming dry gas, mercaptan, acid gas and process gas) from heavy industrial gas, such as coking liquefied gas unit, light hydrocarbon liquefied gas desulfurization tower unit, hydrogenation liquefied gas desulfurization tower unit, acid water stripping unit, rich liquid flash tower unit, acid gas combustion and primary condensation unit.

8. 10 thousand T / a waste liquid desulfurization and sulfuric acid production process: It includes sulfur foam drying, XA sulfur burning and other pretreatment devices, which mainly treat the sulfur foam waste liquid outside the battery limit transported by pipeline. It involves purification system, drying system, conversion section and acid making device, etc., and is mainly used to wash the clean sulfur foam into sulfuric acid product through power wave, acid circulation and heat exchange drying.

9. 4 million T / a indirect coal liquefaction process: It is a process technology in which coal reacts with oxygen and water vapor at high temperature, and then is combined into liquid fuel under the action of catalyst. It involves 9 systems, including Fischer Tropsch synthesis unit, catalyst reduction unit, wax filtration unit, tail gas decarbonization unit, fine desulfurization unit, synthetic water treatment unit, liquid intermediate raw material tank farm unit, low-temperature oil washing unit, deoxygenated water and condensate refining station unit.

10. 1 million T / a hydrocracking process: It is a course of working in which heavy oil is converted into light oil (gasoline, kerosene, diesel oil or feedstock of catalytic cracking and cracking to olefin) by hydrogenation, cracking and isomerization reaction of hydrogen under high pressure and temperature under the action of catalyst in a petroleum refining process. It includes 16 systems, such as feed oil transfer buffer tank, feed oil filter, filtered feed oil buffer tank, hydrogenation feed pump and hydrogen mixing point pipeline.

11. 350 thousand T / a polyethylene production process: With 19 systems, it uses steam cracking ethylene as the main raw material to produce various brands of resins such as injection molding, extrusion and blow molding, especially chlorinated polyethylene special materials.

12. 8 thousand T / a cis polybutadiene rubber process: It adopts solution polymerization method. The raw materials mainly include monomer butadiene and solvent, solvent aliphatic hydrocarbon, alicyclic hydrocarbon, aromatic hydrocarbon and mixed hydrocarbon. The auxiliary materials include catalyst, antioxidant and dispersant. The systems involved include: alkali liquor, aluminum agent and nickel agent preparation system, boron agent high-level system, premixing kettle system, stripping gas condensate buffer tank system and light removal tower system, etc.

13. 200 T / h sour water stripping process: It uses steam stripping method to treat acidic water (an aqueous solution containing volatile weak electrolyte, phenol, cyanide, oil and other pollutants). The systems involved include sour water degassing tank system, sour water stripper system, purified water system at the bottom of sour water stripper, deodorizer absorber system and medium pressure steam system, etc.

14. 500 thousand T / a gas fractionation process: It is aimed at the liquefied gas produced after processing in the refinery. The gas fractionation can separate the liquefied gas into suitable monomer hydrocarbons or alkylated and superimposed raw materials or other high value-added petroleum butane raw materials, such as propylene, n-butene and isobutylene.

The systems involved include: depropanizer feed system, depropanizer reflux system, deethanizer system, propylene rectification column system, purified compressed air system and so on.

15. 800 m$^3$ / h natural gas hydrogen production process: It adopts steam reforming gas generation and pressure swing adsorption purification separation hydrogen extraction. After pressurized desulfurization, natural gas is cracked and reformed with steam in a special reformer filled with catalyst, and then hydrogen is purified by pressure swing adsorption. The systems involved include: natural gas desulfurization, hydrogen reaction and conversion, and intermediate transformation of converted gas.

16. 100 m$^3$ / h formic acid to carbon monoxide process: The decomposition reaction of formic acid and concentrated sulfuric acid is adopted. The formic acid is pumped into the top of the dehydration reactor by the formic acid metering pump to conduct dehydration reaction with sulfuric acid, and the product enters the water washing tower to adjust the pH value to neutral. The procedures involved are: dehydration reaction of formic acid and concentrated sulfuric acid, alkali washing, water washing and pressurization of carbon monoxide, etc.

17. 0.075 T / h ammonium nitrate to nitrous oxide process: It pours ammonium nitrate, a small amount of water and catalyst into the melting pot, heats them, stirs them to form a liquid, and then feeds them into the supply tank through the feeding tank. The procedures involved are: ammonium nitrate melting, nitrous oxide purification, drying, buffering and liquefaction, etc.

18. 500 m$^3$ / h water electrolysis hydrogen production process: In the electrolytic cell, the desalted water from the alkali liquor circulating pump is decomposed under the action of direct current, and hydrogen and oxygen are respectively generated on the cathode and anode plates of the electrolytic cell, and then input to the respective purification devices through pipes for further purification. The procedures are: hydrogen and oxygen washing separation, alkali liquor cooling circulation, etc.

We use HAZOP reports of these processes as the experimental corpus of this paper, which can ensure the universality and universality of the follow-up experiments and can comprehensively measure the effectiveness of our work.

Table 3: Information on hazard data sets.

| Theme | Level #1 | Level #2 | Level #3 | Level #4 | Level #5 |
|---|---|---|---|---|---|
| Possibility | 419 | 1760 | 1607 | 1134 | 949 |
| Severity | 1570 | 2732 | 1353 | 170 | 44 |
| Risk | 2902 | 2577 | 335 | 55 | - |

We perform pre-processing operations such as document sorting and text cleaning on HAZOP reports, and collect 5869 hazards under each theme, see Table 3. We shuffle these hazards and divide them into training set, test set and validation set with a ratio of 8:1:1.

## 5.   EXPERIMENT & ANALYSIS

### 5.1. Experiment setting

In each trial experiment, the main parameters are the same. For example, the size of the pre-trained BERT is base, the optimizer is Adam with a learning rate of 1e-5, the epoch of training is 50, and the batch size is 128. We take the average experimental results of 5 repetitions as the evaluation report. According to the convention, the evaluation metrics are F1, precision, recall, where, F1 is the compromise between precision and recall, which is more representative.

### 5.2. Trial model

Existing models in the similar study [6] are considered as comparison experimental baselines:

(1)   BiLSTM-Attention-FC (base#1): Based on the hazard vector generated by Word2vec, it carries out feature extraction through the combination of BiLSTM model and Attention mechanism, and uses fully connected neural network (FC) to predict the level of hazard.

(2)   BERT-FC (base#2): It uses the hazard vector generated by BERT to directly predict the hazard level through FC.

(3)   BERT-BiLSTM-FC (base#3): It uses the hazard vector generated by BERT and the additional features extracted by BiLSTM to predict the level of hazard through FC.

(4)   BERT-BiLSTM-Attention-FC (base#4): It is similar to base#3, except that it uses the combined feature encoder of BiLSTM and Attention to extract additional features.

Ablation experiments to evaluate the effectiveness of our DLGM. Briefly reiterate that DLGM is a new model with BERT, grey guidance under FSGM(1, 1) and HFFNN:

(5)   DLGM#1: It is a DLGM without grey guidance under FSGM(1, 1), i.e., BERT-HFFNN, which is used to mainly verify the benefits of FSGM(1, 1).

(6)   DLGM#2: It is a DLGM that replaces FSGM(1, 1) with GM(1, 1) to mainly evaluate whether embedding Fourier series into GM(1, 1) can contribute to hazard classification, i.e., BERT-GM(1, 1)-HFFNN.

(7)   DLGM#3: It is a DLGM without HFFNN to mainly check the gain brought by HFFNN, i.e., BERT- FSGM(1, 1)-FC. It also predicts the level of hazard through FC.

(8)   DLGM#4: The proposed complete model, i.e., BERT- FSGM(1, 1)-HFFNN.

### 5.3. Result analysis

Table 4-6 list the total evaluation results of different hazard classifiers under the three themes in turn, where, "test" refers to the test set and "val" refers to the validation set. In addition, for more intuitive analysis, we prepare a series of figures to compare the performance competition between various models (see Appendix and others). There are the following major observations.

Table 4: Performance evaluation results (%) of hazard classification models under the severity theme.

| Model | Precision | | Recall | | F1 | |
|---|---|---|---|---|---|---|
| | test | val | test | val | test | val |
| base#1 | 76.42 | 73.55 | 80.60 | 80.68 | 77.81 | 76.09 |
| base#2 | 80.36 | 79.56 | 84.14 | 81.99 | 81.59 | 80.40 |
| base#3 | 80.47 | 78.48 | 84.65 | 82.19 | 82.06 | 79.79 |
| base#4 | 80.57 | 79.17 | 84.54 | **83.44** | 81.97 | 80.90 |
| DLGM #1 | 81.62 | 80.11 | 85.48 | 83.14 | 83.12 | 81.23 |
| DLGM #2 | 81.48 | 79.17 | 84.15 | **83.44** | 82.53 | 80.90 |
| DLGM #3 | 80.91 | 79.64 | 85.07 | 82.55 | 82.49 | 80.72 |
| DLGM #4 | **81.97** | **80.33** | **85.87** | 83.39 | **83.43** | **81.44** |

Table 5: Performance evaluation results (%) of hazard classification models under the possibility theme.

| Model | Precision | | Recall | | F1 | |
|---|---|---|---|---|---|---|
| | test | val | test | val | test | val |
| base#1 | 67.51 | 65.03 | 65.95 | 66.66 | 66.03 | 65.63 |
| base#2 | 70.18 | 69.00 | 71.12 | 72.00 | 70.60 | 70.19 |
| base#3 | 69.66 | 69.42 | 70.36 | 71.16 | 69.90 | 69.94 |
| base#4 | 70.08 | 69.36 | 70.74 | 72.17 | 70.38 | 70.51 |
| DLGM #1 | **71.48** | **70.30** | 70.68 | 71.96 | 71.01 | 70.90 |
| DLGM #2 | 70.54 | 69.91 | 71.08 | **72.52** | 70.79 | 70.89 |
| DLGM #3 | 70.18 | 69.72 | 71.12 | **72.52** | 70.60 | 70.74 |
| DLGM #4 | 71.39 | 69.94 | **72.09** | **72.52** | **71.72** | **70.93** |

Table 6: Performance evaluation results (%) of hazard classification models under the risk theme.

| Model | Precision | | Recall | | F1 | |
|---|---|---|---|---|---|---|
| | test | val | test | val | test | val |
| base#1 | 66.02 | 72.22 | 69.65 | 73.66 | 67.00 | 72.22 |
| base#2 | 71.68 | 76.35 | 74.30 | 77.70 | 72.81 | 76.67 |
| base#3 | 73.29 | 74.55 | 71.48 | 80.71 | 72.17 | 77.16 |
| base#4 | 73.17 | 75.16 | 74.07 | 79.77 | 73.47 | 77.14 |
| DLGM #1 | 72.57 | 75.69 | 73.97 | 80.83 | 73.20 | 77.76 |
| DLGM #2 | 73.13 | 75.47 | **74.82** | 81.43 | 73.72 | 78.04 |
| DLGM #3 | 72.96 | 76.42 | 74.57 | 80.60 | 73.63 | 77.81 |
| DLGM #4 | **74.03** | **77.42** | 74.74 | **81.73** | **74.33** | **79.25** |

Macroscopically, it can be seen from Table 4-6 and Appendix 1-3 that DLGM# 4 performs better and is remarkable. Its F1 performance under the three themes reaches the best on both the test set and the validation set. In addition, its precision performance and recall performance are also in the lead, which proves effectiveness and progressiveness of our model, and can more reliably distinguish the hazard level. This also reflects the respective effectiveness of grey guidance under FSGM(1, 1) and feature fusion under HFFNN conditioned on the view that hazard is a time series, which is worth looking forward to.

Fig.7 illustrates the F1 performance comparison between DLGM#4 and existing hazard classification models under severity theme, possibility theme and risk theme, where the abscissa represents the performance difference between models and the ordinate represents the pair by pair models for comparison. For example, DLGM#4 − base#1 indicates how much higher the performance of DLGM#4 is than that of base#1. Note that this is also consistent with the following figures. Obviously, the performance of our DLGM#4 model is ahead of the existing models under three themes. Especially for base#1, the performance of DLGM#4 on the test set exceeds that of it by more than 5, 5 and 7 percentage points in turn, and the same is true on the validation set, which indirectly reflects the rationality of using BERT to characterize hazard vectors. For base#2, DLGM#4 leads many of its performance under the three themes, with 1.84 and 1.04 percentage points over the severity theme, 1.12 and 0.74 over the possibility theme, 1.52 percentage points over the risk theme on the test set, and even more than 2 percentage points on the validation set, reaching 2.58, which can reflect that FSGM(1, 1) and HFFNN have the power to boost the model to a certain extent. For base#3 and base#4, DLGM#4 also leads by 2 percentage points at most and 1 percentage point at least, which still shows its more powerful competitiveness.

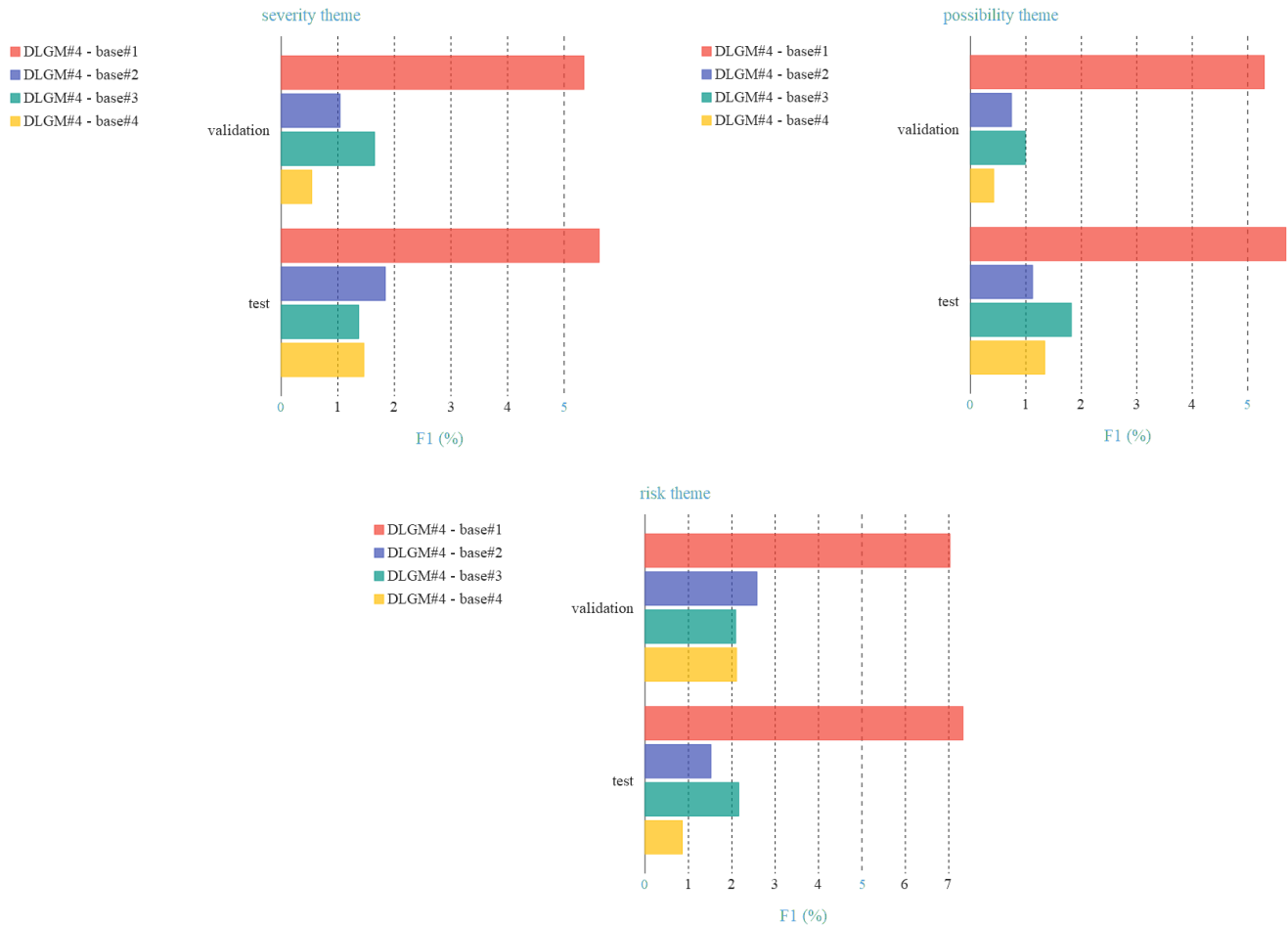

Fig.7: Performance comparison between DLGM#4 and existing hazard classifiers under severity theme, possibility theme and risk theme.

Fig.8 depicts the effect of grey guidance under FSGM(1, 1) for the DLGM classifier, where "P" and "R" respectively represent precision performance and recall performance. It can be observed that the profit of FSGM(1, 1) on the risk theme are commendable, which improves all the performance of the classifier on both the test set and the validation set. Among them, F1 and P on the test set, as well as F1 and R on the validation set, have a gain of more than 1 percentage point. The performance of FSGM(1, 1) on the theme of severity is similar to that on risk, the results of six evaluations also imply the benefits it brings to the DLGM classifier.

Under the possibility theme, although the P performance of the DLGM on the test set and the validation set is slightly weakened, the other four evaluation results are positive, so it is also acceptable on the whole. Undoubtedly, FSGM(1, 1) is feasible and plays an important role for the DLGM classifier, and it has a prospect in the classification of the hazards viewed from the perspective of time series.

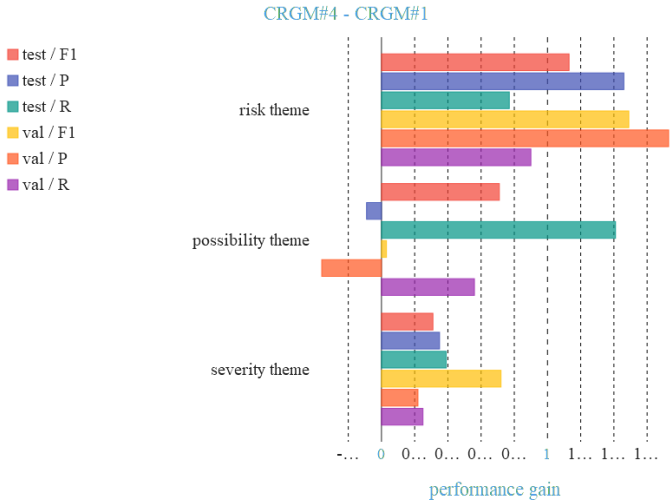

Fig.8: Performance profit evaluation of FSGM(1, 1).

Fig.9 portrays the performance gain of the conceived HFFNN to the proposed DLGM classifier. Evidently, all the evaluation results under the three themes are overflowing with positive attitudes, and nearly half of the evaluation results have increased by about one percentage point. In addition, for both F1 and P performance on the test set, the effect provided by HFFNN is more striking, regardless of which theme. It can be concluded that HFFNN is effective in the process of hazard feature capture and fusion and is irreplaceable for the DLGM classifier.

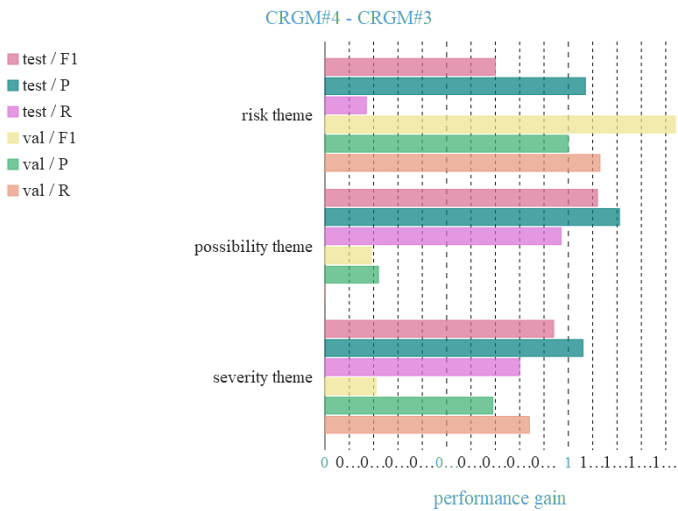

Fig.9: Performance profit evaluation of HFFNN.

Fig.10 is a comparison of whether FSGM(1, 1) is more suitable for hazard classification than GM(1, 1), that is, the feasibility of introducing Fourier series to approximate the fluctuation of hazard time series is judged by the experimental results of hazard classification. It is not difficult to see that FSGM(1, 1) has overwhelming superiority, except for two slight negative exceptions. For some evaluation results, such as the R performance on the validation set under the severity theme and the P performance on the validation set under the risk theme, the gain brought by FSGM(1, 1) is even more than one percentage point higher than that brought by GM(1, 1). Undeniably, the evaluation results prove that it is feasible and effective to embed Fourier series into GM (1, 1) for hazard classification.

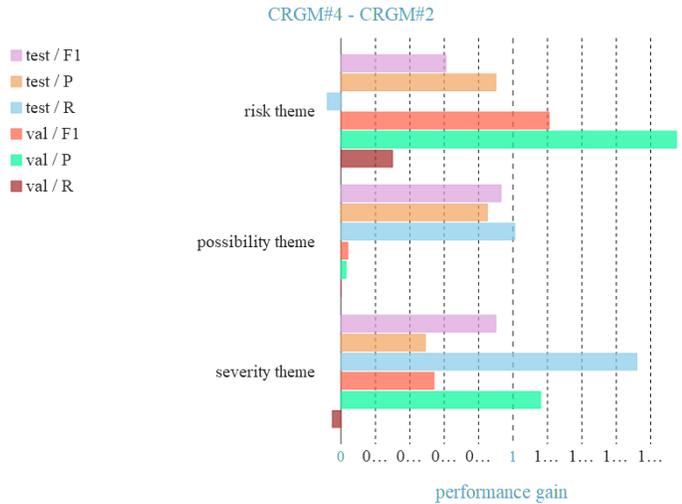

Fig.10: Performance comparison of FSGM(1, 1) and GM(1, 1).

To sum up, our DLGM, a new hazard classifier established by HFFNN under the guidance of FSGM(1, 1), has promising and gratifying aptitudes. FSGM(1, 1) and HFFNN can ameliorate the hazard classification, which is reliable and effective. We hope our research can contribute added value to the daily practice in industrial safety and provide support for the pioneering of hazard classification.

## 6. DISCUSSION

Encouragingly, the proposed DLGM is instrumental in the classification of hazards, which is more competitive and promising than the previous models. What makes sense is that this may be a feedback reward what it can fuse is multi-directional information. One is the gray guidance, the other is the feature screening and fusion about sentence level features and multi-local features. Besides, it fuses the semantic information of hazard in vectorization, which is also a kind of incentive. The multi-directional information fusion can elevate the DLGM classifier to understand the hazard more fully.

Some limitations of this paper need to be noted, although they are irrelevant. It should be reminded that the hazard analyzed by HAZOP can be naturally regarded as a class of time series due to its unique attributes. Thus, this paper has the opportunity to train a more effective hazard classifier under the guidance of grey model modeling hazards, and the guidance here can be considered as a class of potential feature to assist in deep learning training. It is worth noting that this paper is aimed at the classification of the hazards, not the grey model itself. How the grey model fits the trend of the hazard itself is of very weak value to this study, so it is unnecessary to conduct error analysis and so on. The final experimental results of hazard classification are sufficient to evaluate whether they are effective for this study. In addition, the grey model itself is a very large and systematic research field, and the consideration of other types of grey models is beyond the scope of this paper. Another reminder is that No Free Lunch Theorem [22] tells us that no algorithm can be applied to all real problems. Similarly, this paper also faces the challenge of how to be applied to other classification tasks.

In short, our research is profound and meaningful, we hope that it can inspire other relevant researchers. For the limitations, we expect that there will be other follow-up studies to focus on them in the similar way to the development of grey models and neural networks [58].

Future work is mainly committed to providing migration and reference for the safety analysis of other types of processes, since our DLGM classifier is trained on 18 processes, mainly petroleum type, gas type, coal and sulfur type, and fails to serve and interface with other types of industrial processes, such as aviation related processes and concrete related processes. So, we need to strengthen and harmonize the audience size and popularity of hazard classifier. Sincerely, we hope that our research can burst out its due value and serve the intelligent progress of industrial safety more comprehensively and orderly.

## 7.  CONCLUSION

The national policy and the realistic situation force HAZOP to carry out safety analysis on industrial processes, especially for China where the industrialization is developing orderly. In this paper, we propose a novel hazard classifier, called DLGM, which can distinguish the hazards analyzed by HAZOP and provide decision support for experts, engineers and employees. DLGM measures hazards from the severity theme, possibility theme and risk theme with respective levels. To our knowledge, our work is the first to conduct such a comprehensive research on hazard classification.

DLGM fuses multi-directional information. One is the semantic information in the hazard vector formed by BERT, the other is gray guidance in the sense of structural parameters modeled by FSGM(1, 1) grey model, and the last is the feature capture, screening and fusion about sentence level features and multi-local features controlled by HFFNN. Where, the ideas involved include that the hazard vector can be regarded as a kind of time series (HTS), the introduction of Fourier series into GM(1, 1) is considered to approximate the forcing of HTS, and the processing of features by HFFNN can strengthen the classifier to recognize the hazard.

DLGM is built on the deep learning architecture. Specifically, we first transform the hazard into the HTS through BERT, then conceive the FSGM(1, 1) and inject its grey guidance into HTS to form HTSGG, and finally design the HFFNN to process HTSGG and classify it by fully connected neural network.

We hold 18 industrial processes as application cases, such as 1.2 million T / a heavy oil catalytic cracking process and 500 $m^3$ / h water electrolysis hydrogen production process, etc. On this basis, we launch a series of experiments. The competition experiment implies the effectiveness and progressiveness of DLGM, and the ablation experiment reflects the feasibility of the guidance of FSGM(1, 1), and the advancement of HFFNN. We hope our work can supplement additional incentives and added value for the daily practice of industrial safety, as well as enlighten researchers committed to hazard classification, and stabilize the orderly development of national industry.

## DECLARATION OF COMPETING INTEREST

The authors declare that they have no known competing financial interests or personal relationships that could have appeared to influence the work reported in this paper.

## ACKNOWLEDEMENTS

## REFERENCE

[1] China General Administration of work safety. (July 2013). Guidance on strengthening safety management of chemical process, 31. www.gov.cn/gongbao/content/2013/content_25197 13.htm.

[2] Emergency Management Department of the people's Republic of China. (July 2019). Management measures for emergency management standardization. https://www.mem.gov.cn/gk/tzgg/ tz/201907/t20190707_321229.shtml.

[3] Emergency Management Department of the people's Republic of China. (November 2020). Catalogue of safety classification and rectification of hazardous chemical enterprises. https://www.mem.gov.cn/gk/tzgg/tz/202011/t20201103_371291.shtml. Accessed 2022.

[4] Emergency Management Department of the people's Republic of China. HAZOP. https://www.mem.gov.cn/.

[5] Macasaet, D. , Bandala, A. , Illahi, A. A. , Dadios, E. , & Lauguico, S. . (2020). Hazard Classification of Toluene, Methane and Carbon Dioxide for Bomb Detection Using Fuzzy Logic. 2019 IEEE 11th International Conference on Humanoid, Nanotechnology, Information Technology, Communication and Control, Environment, and Management ( HNICEM ). IEEE.

[6] Feng, X., Dai, Y., Ji, X., Zhou, L., & Dang, Y. (2021). Application of natural language processing in HAZOP reports. Process Safety and Environmental Protection, 155, 41-48.

[7] Wang, B. , Wu, C. , Kang, L. , Reniers, G. , & Huang, L. . (2018). Work safety in china's thirteenth five-year plan period (2016–2020): current status, new challenges and future tasks. Safety Science, 104, 164-178.

[8]   Wang, B, Wu, C. , Reniers, G. , Huang, L. , Kang, L. , & Zhang], L. . (2018). The future of hazardous chemical safety in china: opportunities, problems, challenges and tasks. Science of The Total Environment.

[9]   Chen, C. , & Reniers, G. . (2020). Chemical industry in china: the current status, safety problems, and pathways for future sustainable development. Safety Science, 128.

[10]  Wang, Z., Zhang, B., & Gao, D. (2022). A novel knowledge graph development for industry design: A case study on indirect coal liquefaction process. Computers in Industry, 139, 103647.

[11]  Standard, B., & IEC61882, B. S. (2001). Hazard and operability studies (HAZOP studies)-Application guide. International Electrotechnical Commission.

[12]  Dunjo, J. , Fthenakis, V. , Vilchez, J. A. , & Arnaldos, J. . (2010). Hazard and operability (hazop) analysis. a literature review. Journal of Hazardous Materials, 173(1-3), 19-32.

[13]  Zhao, Y., Zhang, B., & Gao, D. (2022). Construction of petrochemical knowledge graph based on deep learning. Journal of Loss Prevention in the Process Industries, 104736.

[14]  Wang, Z., Zhang, B., & Gao, D. (2021). Text Mining of Hazard and Operability Analysis Reports Based on Active Learning. Processes, 9(7), 1178.

[15]  Peng, L., Gao, D., & Bai, Y. (2021). A Study on Standardization of Security Evaluation Information for Chemical Processes Based on Deep Learning. Processes, 9(5), 832.

[16]  Gordon, L. B., Cartelli, L., & Graham, N. (2018). A complete electrical shock hazard classification system and its application. IEEE Transactions on Industry Applications, 54(6), 6554-6565.

[17]  Ouyang, L., Che, Y., Yan, L., & Park, C. (2022). Multiple perspectives on analyzing risk factors in FMEA. Computers in Industry, 141, 103712.

[18]  Jahani, A. (2019). Sycamore failure hazard classification model (SFHCM): an environmental decision support system (EDSS) in urban green spaces. International journal of environmental science and technology, 16(2), 955-964.

[19]  Fang, W., Luo, H., Xu, S., Love, P. E., Lu, Z., & Ye, C. (2020). Automated text classification of near-misses from safety reports: An improved deep learning approach. Advanced Engineering Informatics, 44, 101060.

[20]  Zhang, R., Jia, C., & Wang, J. (2022). Text emotion classification system based on multifractal methods. Chaos, Solitons & Fractals, 156, 111867.

[21]  Gu, A., Gulcehre, C., Paine, T. L., Hoffman, M., & Pascanu, R. (2020). Improving the Gating Mechanism of Recurrent Neural Networks. Proceedings of the 37th International Conference on Machine Learning, PMLR. 119:3800-3809.

[22]  Ho, Y. C., & Pepyne, D. L. (2002). Simple explanation of the no-free-lunch theorem and its implications. Journal of optimization theory and applications, 115(3), 549-570.

[23]  Ferreira, C. , Ribeiro, J. , & Freire, F. . (2019). A hazard classification system based on incorporation of reach regulation thresholds in the usetox method. Journal of Cleaner Production, 228(AUG.10), 856-866.

[24] Barber, S. , Boulinquiez, M. , Caprio, E. D. , Candeal, J. , & Kernen, H. . (2017). Hazard classification and labelling of petroleum substances in the european economic area - 2017. CONCAWE Reports(13), 1-317.

[25] Stiernstroem, S. , Wik, O. , & Bendz, D. . (2016). Evaluation of frameworks for ecotoxicological hazard classification of waste. Waste Management, 58(dec.), 14-24.

[26] Stein, R., Jaques, P., & Valiati, J. (2019). An analysis of hierarchical text classification using word embeddings. Information Sciences, 471, 216-232.

[27] Kong, L., Li, C., Ge, J., Zhang, F., Feng, Y., Li, Z., & Luo, B. (2020). Leveraging multiple features for document sentiment classification. Information Sciences, 518, 39-55.

[28] Zhan, Z. , Zhou, J. , & Xu, B. . (2022). Fabric defect classification using prototypical network of few-shot learning algorithm. Computers in Industry, 138, 103628.

[29] Liu, X., Yao, L., Tomotake, F., Soji, Y., Wentai, Z., Amit, R., Levent, K., & Kenji, S. (2022). Graph neural network-enabled manufacturing method classification from engineering drawings. Computers in Industry, 142, 103697.

[30] Wen, W., Hua, G., Wen, D., Zhe, X., & Xin, Ren. (2022). ABL-TC: A lightweight design for network traffic classification empowered by deep learning, Neurocomputing, 489, 333-344.

[31] Dai, W. , Li, D. , Tang, D. , Wang, H. , & Peng, Y. . (2022). Deep learning approach for defective spot welds classification using small and class-imbalanced datasets. Neurocomputing, 477, 46-60.

[32] Prakash, S., Agrawal, A., Singh, R., Singh, R.K. and Zindani, D. (2022). A decade of grey systems: theory and application – bibliometric overview and future research directions. Grey Systems: Theory and Application, Vol. ahead-of-print No. ahead-of-print. https://doi.org/10.1108/GS-03-2022-0030.

[33] Wang, Q. , & Song, X. . (2019). Forecasting china's oil consumption: a comparison of novel nonlinear-dynamic grey model (GM), linear gm, nonlinear GM and metabolism GM. Energy, 183(Sep.15), 160-171.

[34] Muya, B. , Ab, A. , & Ea, A. . A modified GM(1,1) model to accurately predict wind speed. (2021). Sustainable Energy Technologies and Assessments, 43, 100905.

[35] Wang, M. , Wu, L. , & Guo, X. . (2022). Application of grey model in influencing factors analysis and trend prediction of carbon emission in Shanxi province. Environmental Monitoring and Assessment,194, 542.

[36] Fattor, M. V. , & Vieira, M. A. . (2019). Application of human HAZOP technique adapted to identify risks in brazilian waste pickers' cooperatives. Journal of Environmental Management, 246(SEP.15), 247-258.

[37] Lim, C. H. , Lim, S. , Bing, S. H. , Ng, W. , & Lam, H. L. . (2021). A review of industry 4.0 revolution potential in a sustainable and renewable palm oil industry: HAZOP approach. Renewable and Sustainable Energy Reviews, 135, 110223.

[38] Lim, C. H. , Lam, H. L. , & Pei Qin, W. N. . (2018). A novel HAZOP approach for literature review on biomass supply chain optimisation model. Energy. 146, 13-25.

[39] Wang, W. , Qin, J. , Yu, M. , Li, T. , & Li, J. . (2019). A reliability analysis of cfetr csmc heat treatment system based on RPN-HAZOP method. IEEE Transactions on Plasma Science, PP(99), 1-5.

[40] Patle, D. S. , Agrawal, V. , Sharma, S. , & Rangaiah, G. P. . (2021). Plantwide control and process safety of formic acid process having a reactive dividing-wall column and three material recycles. Computers & Chemical Engineering, 147(2), 107248.

[41] Hrjs, A. , Sn, B. , Ah, A. , Hm, C. , & Mk, D. (2022). A fuzzy-HAZOP/ant colony system methodology to identify combined fire, explosion, and toxic release risk in the process industries. Expert Systems with Applications, 192, 116418.

[42] Marhavilas, P. K. , Filippidis, M. , Koulinas, G. K. , & Koulouriotis, D. E. . (2021). Safety-assessment by hybridizing the mcdm/ahp & HAZOP-DMRA techniques through safety's level colored maps: implementation in a petrochemical industry-sciencedirect. Alexandria Engineering Journal. 61, 9, 6959-6977.

[43] Cheraghi, M. , Baladeh, A. E. , & Khakzad, N. . (2019). A fuzzy multi-attribute HAZOP technique (FMA-HAZOP): application to gas wellhead facilities. Safety Science, 114, 12-22.

[44] Zheng, Z. , Lu, X. Z. , Chen, K. Y. , Zhou, Y. C. , & Lin, J. R. (2022). Pretrained domain-specific language model for general information retrieval tasks in the AEC domain. Computers in Industry, 142, 103733.

[45] Nicola M., Irlan G., Gualtiero F. (2022). Enhancing Industry 4.0 standards interoperability via knowledge graphs with natural language processing. Computers in Industry, 140, 103676.

[46] Donghwa K., Pilsung K. (2022). Cross-modal distillation with audio–text fusion for fine-grained emotion classification using BERT and Wav2vec 2.0. Neurocomputing, 506, 168-183.

[47] Ambalavanan, A. K. , & Devarakonda, M. V. . (2020). Using the contextual language model bert for multi-criteria classification of scientific articles. Journal of Biomedical Informatics, 112(3), 103578.

[48] Kayacan, E. , Ulutas, B. , & Kaynak, O. . (2010). Grey system theory-based models in time series prediction. Expert Systems with Applications, 37(2), 1784-1789.

[49] Wang, X., Xie, N., Yang L. (2022). A flexible grey Fourier model based on integral matching for forecasting seasonal PM2.5 time series.Chaos, Solitons & Fractals, 162, 112417.

[50] Bhattacharyya, A. , Singh, L. , & Pachori, R. B. . (2018). Fourier-bessel series expansion based empirical wavelet transform for analysis of non-stationary signals. Digital Signal Processing, S1051200418300642.

[51] Yao, W. , Weng, Y. , & Catchmark, J. M. . (2020). Improved cellulose x-ray diffraction analysis using fourier series modeling. Cellulose, 27(10), 5563-5579.

[52] Caruso, A. , Bassetto, M. , Mengali, G. , & Quarta, A. A. . (2019). Optimal solar sail trajectory approximation with finite fourier series. Advances in Space Research, 67(9), 2834-2843.

[53] Zhang, Y. Y. , Zang, S. Y. , Zhao, L. , Ma, D. L. , Lin, Y. , & Li, H. . (2022). Estimation of permafrost thermal behavior using fourier series model. Journal of Mountain Science, 19(3), 715-725.

[54] Xu, D. , Tian, Z. , Lai, R. , Kong, X. , & Shi, W. . (2020). Deep learning based emotion analysis of microblog texts. Information Fusion, 64, 1-11.

[55] Muñoz, S., Iglesias, C. (2022). A text classification approach to detect psychological stress combining a lexicon-based feature framework with distributional representations. Information Processing & Management, 59, 5, 103011.

[56] Wang, Z., Liu, H., Liu, F., & Gao, D. (2022). Why KDAC? A general activation function for knowledge discovery. Neurocomputing, 501, 343-358.

[57] Liu, Y., Wang, L., Li, H. & Chen, X. (2022). Multi-focus image fusion with deep residual learning and focus property detection. Information Fusion, 86–87, 1-16.

[58] Samek, W. , Montavon, G. , Lapuschkin, S. , Anders, C. J. , & Muller, K. R. . (2021). Explaining deep neural networks and beyond: a review of methods and applications. Proceedings of the IEEE, 109, 3, 247-278.

*APPENDIX*

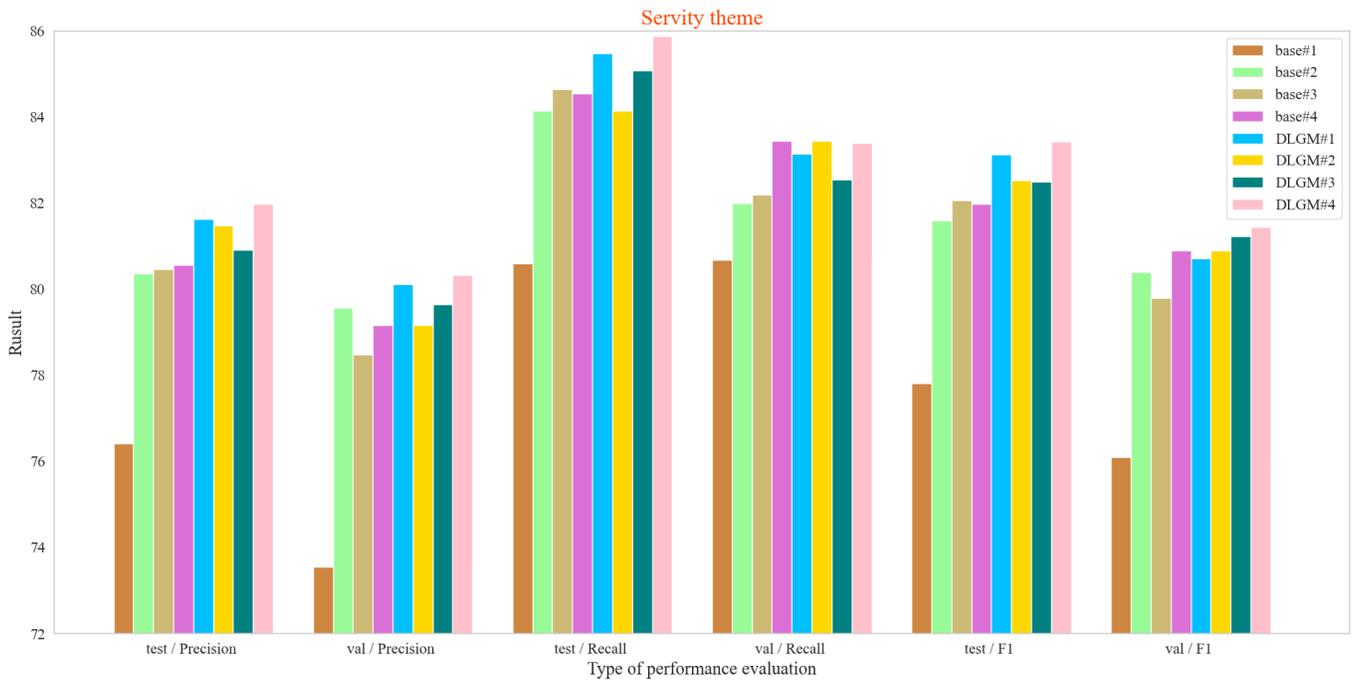

Appendix 1: Performance evaluation results (%) of hazard classifiers under the severity theme.

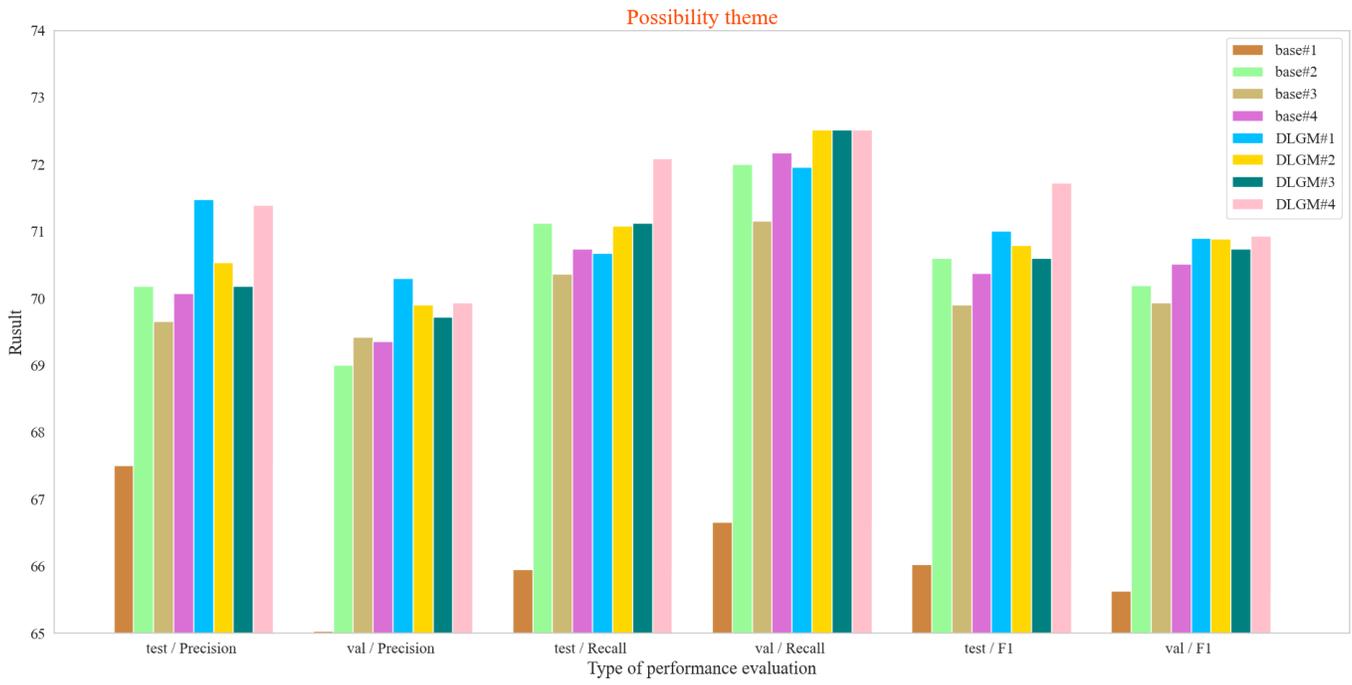

Appendix 2: Performance evaluation results (%) of hazard classifiers under the possibility theme.

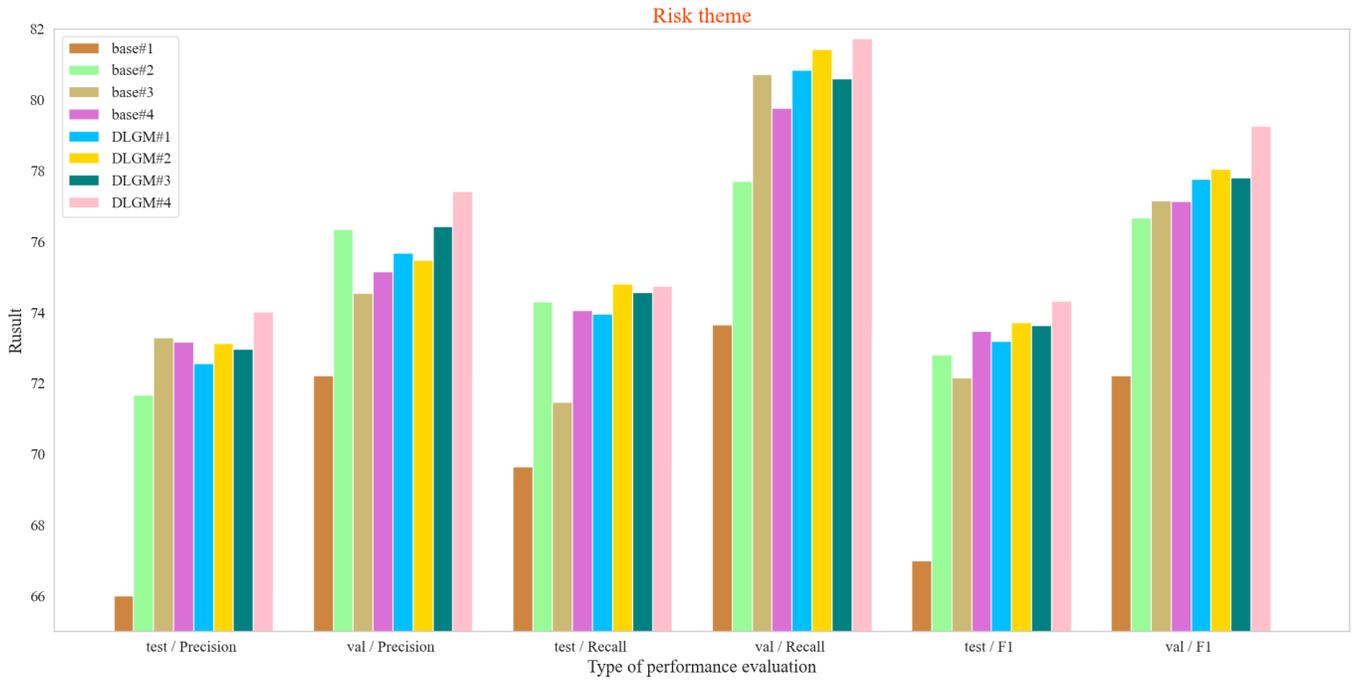

Appendix 3: Performance evaluation results (%) of hazard classifiers under the risk theme.